\documentclass[12pt,a4paper]{scrartcl}

\DeclareOldFontCommand{\bf}{\normalfont\bfseries}{\mathbf}

\usepackage[utf8]{inputenc}

\usepackage{cite}

\addtokomafont{disposition}{\rmfamily}
\usepackage{newpxtext,newpxmath}
\usepackage{multirow}

\usepackage{subcaption}

\addtokomafont{disposition}{\rmfamily}

\usepackage{nameref}

\usepackage[right]{lineno}

\usepackage{microtype}

\usepackage{changepage}

\usepackage[aboveskip=1pt,labelfont=bf,labelsep=period,singlelinecheck=off]{caption}

\makeatletter
\renewcommand{\@biblabel}[1]{\quad#1.}
\makeatother

\usepackage{color}

\definecolor{Gray}{gray}{.25}

\usepackage{graphicx}

\usepackage{sidecap}

\usepackage{wrapfig}
\usepackage[pscoord]{eso-pic}
\usepackage[fulladjust]{marginnote}
\reversemarginpar

\usepackage{url,lineno,microtype,subcaption}
\usepackage{textcomp}
\usepackage[onehalfspacing]{setspace}
\usepackage[english]{babel}

\usepackage[square,sort,comma,numbers]{natbib}

\usepackage{graphicx}
\usepackage{subcaption}
\usepackage{floatrow}
\usepackage{textcomp}
\usepackage{color,soul}

\usepackage{hyperref}
\usepackage[capitalize, noabbrev]{cleveref}
\hypersetup{
     colorlinks   = true,
     citecolor    = green
}

\begin{document}

\begin{flushleft}
{\Large
\textbf\newline{Socially competent robots: adaptation improves leadership performance in groups of live fish}
}
\newline
Tim Landgraf\textsuperscript{1*},
Hauke J. Moenck\textsuperscript{1},
Gregor H.W. Gebhardt\textsuperscript{1,2},
Nils Weimar\textsuperscript{3},
Mathis Hocke\textsuperscript{1},
Moritz Maxeiner\textsuperscript{1},
Lea Musiolek\textsuperscript{4},
Jens Krause\textsuperscript{3,5},
David Bierbach\textsuperscript{3}
\\
\bigskip
\bf{1} Department of Mathematics and Computer Science, Freie Universität Berlin, Berlin, Germany \\
\bf{2} Computational Systems Neuroscience, Institute of Zoology, University of Cologne, Cologne, Germany \\
\bf{3} Faculty of Life Sciences, Albrecht Daniel Thaer Institute of Agricultural and Horticultural Sciences, Humboldt Universität zu Berlin, Berlin, Germany \\
\bf{4} Department of Computer Science, Humboldt-Universit\"at zu Berlin, Berlin, Germany \\
\bf{5} Leibniz-Institute of Freshwater Ecology and Inland Fisheries, Berlin, Germany
\bigskip
\\
* corresponding author: tim.landgraf@fu-berlin.de

\end{flushleft}

\section*{Abstract}
Collective motion is commonly modeled with simple interaction rules between agents. Yet in nature, numerous observables vary within and between individuals and it remains largely unknown how animals respond to this variability, and how much of it may be the result of social responses. Here, we hypothesize that Guppies (\textit{Poecilia reticulata}) respond to avoidance behaviors of their shoal mates and that "socially competent" responses allow them to be more effective leaders. We test this hypothesis in an experimental setting in which a robotic Guppy, called RoboFish, is programmed to adapt to avoidance reactions of its live interaction partner. We compare the leadership performance between socially competent robots and two non-competent control behaviors and find that 1) behavioral variability itself appears attractive and that socially competent robots are better leaders that 2) require fewer approach attempts to 3) elicit longer average following behavior than non-competent agents. This work provides evidence that social responsiveness to avoidance reactions plays a role in the social dynamics of guppies. We showcase how social responsiveness can be modeled and tested directly embedded in a living animal model using adaptive, interactive robots. 

\linenumbers

\flushbottom

%
\thispagestyle{empty}

\section*{Introduction}
In complex social systems, the dynamics of individual interactions underlie the emergent phenomena on the group level. To reproduce the coordinated motion patterns of shoals and flocks, for example, simple inter-individual rules of attraction and repulsion have been shown to be a sufficient mathematical model of individual behavior \cite{couzin_collective_2002}. Collectives in nature often exhibit substantial phenotypical variation within and between individuals and these factors affect how animals interact. Differences in body size \cite{romenskyy_body_2017,hemelrijk_density_2004}, personality \cite{jolles_dominance_2013,nakayama_experience_2013,kurvers_personality_2009,harcourt_personality_2009,worm_evidence_2018,jolles_consistent_2017} or physiological states \cite{bumann_front_1993,killen_role_2017}, for example, have been shown to predict how individuals behave in social contexts.

In contrast to most computational models of collective behavior, interaction rules in biological systems seem to respond dynamically to different sources of variation. In Sticklebacks (\textit{Gasterosteus aculeatus}), individual differences have been shown to reinforce leader and follower roles \cite{harcourt_personality_2009,nakayama_experience_2013}, highlighting the importance of group composition. Interaction rules may also change over time as a result of increased familiarity between individuals \cite{dugatkin_female_1993,lachlan_who_1998,swaney_familiarity_2001,bierbach_male_2011}. From an ecological point of view, individuals can optimize their success by adjusting the interaction rules in response to their social environment. This eventually leads to a higher Darwinian fitness compared to those that do not adapt, or do so only poorly. Such an ability has been termed ‘social competence’ or ‘social responsiveness’ \cite{taborsky_social_2012,wolf_adaptive_2013}. 


For a fitness-relevant task, for example leadership \cite{strandburg-peshkin_inferring_2018}, we hypothesize that a "socially competent" leader should be more effective than a non-competent conspecific. But which observations does the social competent leader integrate into which behavioral response?

Due to the recursive dynamics of collective systems, this question can not be investigated through observation only. Modeling mathematically how interaction rules change in socially competent agents, and validating these models’ predictions against real-world data is hard for similar reasons, because we can not disentangle which behavioral variation exists independently of the social dynamics and which is the result of a response. Robots that mimic conspecifics are increasingly used to investigate social behavior \cite{krause_interactive_2011}. With robots we have full control over one of the interaction partners, and with that over the existence and properties of social feedback loops. We can, for example, embody models of social competence in a robotic agent and compare its performance to that of a non-competent control behavior.

\begin{figure}
    \centering

    \includegraphics[width=1\textwidth]{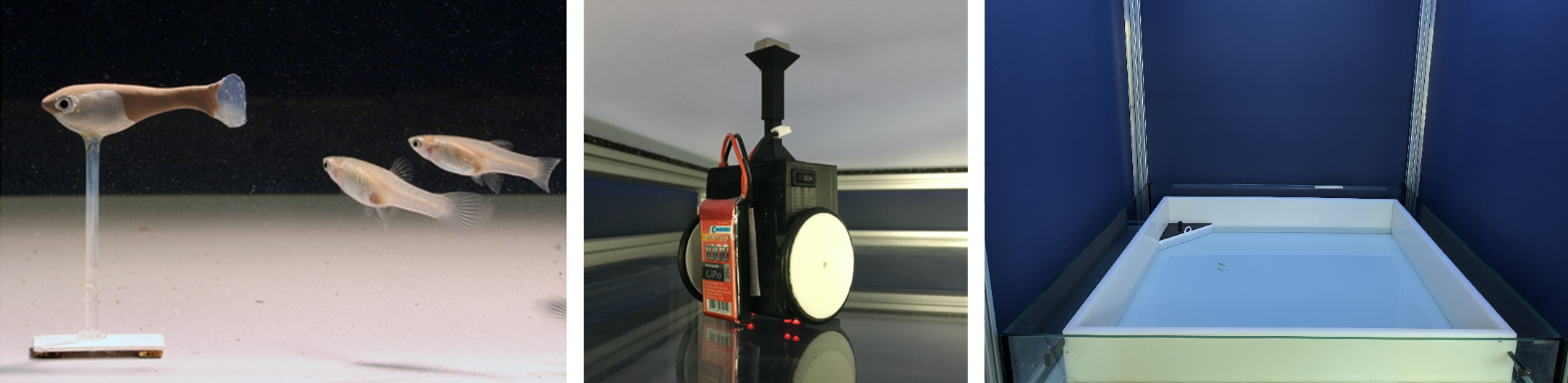}
    \caption{The RoboFish system. The 3-D printed fish replica is attached to a magnetic base plate (left panel). Its movements are controlled by a two-wheeled robot below the fish tank carrying a neodymium magnet (middle panel). The tank is a quadratic (1 m x 1 m) with a triangular start box used as shelter for the live fish at the beginning of a test trial (right panel). The robot control software tracks the position and orientation of both the live fish and the robot in real-time (see \hyperref[sec:methods]{Methods}).}
    \label{fig:robofish_design}
\end{figure}

We developed an interactive robotic guppy (‘RoboFish’ \cite{landgraf_robofish_2016}) that can observe and memorize the interaction partner’s past responses towards its own actions and adjust its interaction rules as a function of these observations (see Figure \ref{fig:robofish_design}). 

Motivated by the fact that most leadership interactions happen in close proximity (see SI\ref{sec:follow_iid}), we have implemented two behavioral subroutines, an ‘approach phase’ in which the robot closes in on a live fish, and a 'lead phase' in which it swims ahead of the fish, along the tank walls as long as the fish stays close (cf. Figure~\ref{fig:robofish_design}).

In many examples of fission-fusion dynamics\cite{aureli_fissionfusion_2008,couzin_behavioral_2006,krause_living_2002} in which animals switch between social and solitary periods (for guppies, see \cite{wilson_network_2013,wilson_dynamic_2014}), animals may respond aversively to social proximity. Such avoidance behavior has been described in guppies and other members of the family Poeciliidae for various types of social contexts such as mating \cite{plath_male_2008,magurran_sexual_2011}, cannibalism \cite{chapman_early_2008}, disease prevention \cite{stephenson_transmission_2018,croft_effect_2011}, or aggressive encounters \cite{bierbach_casanovas_2013}. An avoidance reaction may inform the approaching fish that the approached individual is unwilling to engage in social interactions and a perfect candidate for a behaviorally relevant observation. Here, we defined the socially competent leader as an individual, who detects avoidance reactions and appropriately adjusts its follow-up interaction by approaching more carefully.



The robot quantifies avoidance motions and continuously integrates these measurements into a scalar variable $a_t$ (coined `carefulness') that represents a short term memory of past observations. This variable then defines the angle and speed of the approach: fish that frequently avoid the competent robot produce carefulness values $a_t \approx 1$ resulting in subsequent approaches performed indirectly (at a $\approx90$\textdegree angle) and slowly (at 8 $cm^{-s}$). Observations of no or weak avoidance decrease the carefulness value over time. At the other end of the carefulness spectrum fish are approached with high velocity and directness (maximum of 30 $cm^{-s}$ and $0$\textdegree\ for $a_t = 0$, see \hyperref[sec:methods]{Methods} for details). Note that, in contrast to a fixed mapping, the behavioral observations define direction and magnitude of a change of the carefulness variable. This way, the robot can adapt to the optimal directness and speed a given individual allows.  

If the fish accepts the robot’s approach and stays in proximity ($<$ 12 $cm$ distance) for 2 s, the robot switches to lead phase, swimming along the tank walls as long as the fish stays close ($<$ 28 $cm$ distance with a 1 s tolerance). If the fish falls back, RoboFish switches back to approach phase.

We implemented two variants of a non-competent robot, one that either always uses the same choice of carefulness for its approaches (fixed mode, experiment 1) or one that uses a randomly chosen carefulness value (random mode, experiment 2). In pre-trials, we obtained the distribution of carefulness values for a competent robot. The mean carefulness was used in fixed mode ($\bar{a}=0.528$, see \hyperref[sec:methods]{Methods}) resulting in approaches with moderate speed and directness ($v=19$ $cm^{-s}$ and $\alpha=47$\textdegree). In random mode, the carefulness values were drawn from the reference distribution such that after each trial the distributions matched approximately the social competent reference. To quantify leadership performance, we determined the mean duration the fish followed the robot (total duration of all following episodes divided by their count), the number of approaches RoboFish performed for a given duration of following episodes (the fewer, the better) and the mean avoidance the fish showed throughout the trial. We predicted that a socially competent RoboFish produces less avoidance, is more efficient and elicits longer following episodes than the non-competent controls.

\section*{Results}
We ran a total of 86 trials (46 in experiment 1 and 40 in experiment 2). Over all trials, we observed sustained interest in the robot with a few exceptions of fish that showed pronounced avoidance reactions and no following behavior whatsoever. Pooling all treatments, we observed following behavior totalling to 3.9 hours of theoretically possible 14.3 hours (86 trials of 10 min duration). More than half of all following episodes occur within the first three minutes (104 / 197), accounting for 74 \% of the combined following durations (173 min / 235 min). 

\subsection*{Lower or similar avoidance in socially competent robots}
\begin{figure}[htbp]
    \includegraphics[width=1.1\textwidth]{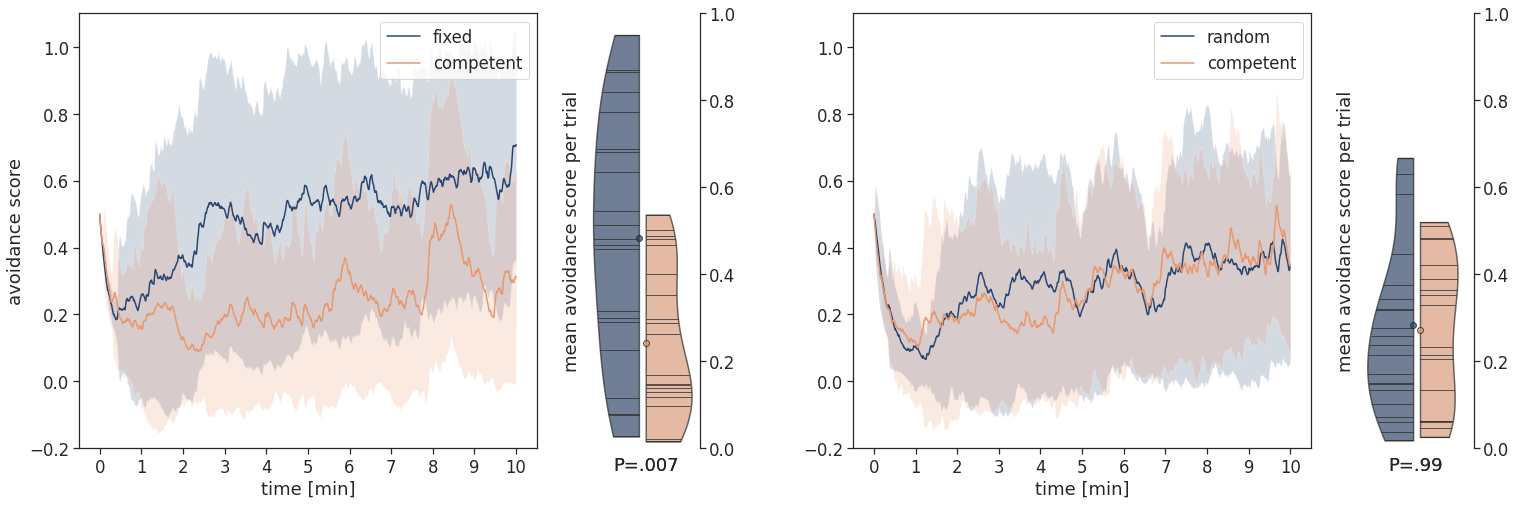}
    \caption{Avoidance scores over time. We quantified fish motions away from the robot when they are sufficiently close (see \hyperref[sec:methods]{Methods}). We tracked this avoidance score over time for both the non-competent modes (blue, line shows the mean avoidance over time, shaded region depicts the 68 \% confidence interval) and the socially competent mode (orange). In experiment 1 (left panel), where the non-competent robot always uses the same carefulness value (fixed mode), the avoidance scores are significantly lower for the socially competent mode. In experiment 2 (right panel), we find no significant differences. Note that we observe a drop in avoidance scores over all settings reflecting the consistent initial interest in the robot.}
    \label{fig:avoidance_results}
\end{figure}

Most fish were attracted by the robot at the beginning of the trial, hence, we consistently observed decreasing mean avoidance scores over time (Figure \ref{fig:avoidance_results}) for both competent and non-competent treatments. While the socially competent robot had a similar per-trial mean carefulness compared to fixed mode (median: 0.53/0.59, N1/2=21/21 U=210, P=.8, CLES=0.52), the avoidance scores were found to be significantly smaller (reporting median [min max]; fixed: 0.64 [0.063 0.96], competent: 0.46	[0.18 0.86], N1/2=23, U=364 P=.03, CLES=.69, Figure \ref{fig:avoidance_results}). 

Comparing to random mode, the socially competent robot had a lower carefulness (random: 0.69 [0.46 to 0.89], competent: 0.59 [0.15 0.89], U=253, P=.14, CLES=.64) and produced higher median avoidance scores (random: 0.33 [0.045 0.72], competent: 0.48	[0.17 0.71], N1/2=22/18, U=133, P=.08, CLES=.66, Figure \ref{fig:avoidance_results}). 

We find higher mean motion speeds of both robot and fish in experiment 1 for the fixed treatment (fixed: 7.19 [2.71 8.65] $cm^{-s}$, competent: 4.99 [2.17 7.78] $cm^{-s}$, U=361, P<.001, CLES=.82), but no such differences in experiment 2 (random: 4.2 [2.01 7.68] $cm^{-s}$, competent: 4.74 [2.13 8.39] $cm^{-s}$, U=160, P=.31, CLES=.6). See also SI\ref{suppl:speed_comparison} for details.

\subsection*{Fish follow socially competent robots longer}
In both experiments, the socially competent robot evoked longer mean follow episodes and longer total following durations. 

\begin{figure}[htbp]
    \includegraphics[width=\textwidth]{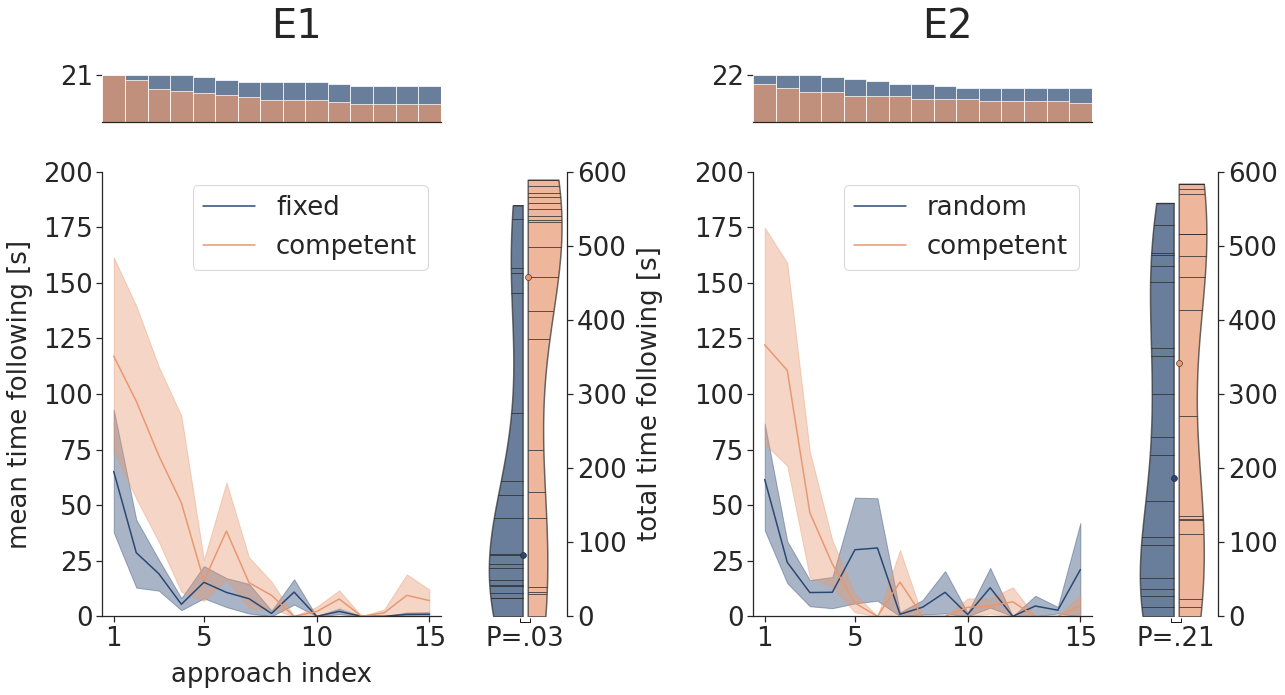}
    \caption{Comparison of follow episode durations. We compare the follow episode durations within a trial for socially competent robots (orange) and non-competent robots (blue) for both main experiments (columns E1 and E2). Each panel shows the mean duration of follow episodes over the approach index (i.e. a sequential ID of approach phases, left sub-panel). The approach phase count is shown in the bar plot above the left sub-panels. For the sake of clarity, the plot has been cut to include only the first 15 approaches. See SI for complete plots.  The distribution of the total following durations (right sub-panel) in the half-violin plots. Median values are depicted with a orange and blue circle, respectively. P-values of a Mann-Whitney U-test are given under the violin plot. Long follow episodes are predominantly initiated at the beginning of the trial, after the first few approaches. Differences between treatments pertain to the first 5 approaches in which the socially competent robots perform considerably better.}
    \label{fig:sum_follow_durations}
\end{figure}

In experiment 1 we observed a pronounced difference of the per-trial mean following duration (competent: 53.4 s [0 s to 589 s] in competent mode and 3.1 s [0 s to 138.6 s] in fixed mode,U=124, P=.016, CLES=.71, see SI\ref{suppl:mean_follow}). 

In experiment 2, although less pronounced, we observed higher mean follow episode durations for the socially competent robot (random: 7.2 s [0 s 157.5s], competent: 24.4 s [0 s 583.5 s], U=147, P=.17,CLES=.62). 

In both treatments the majority of live fish followed at the beginning of a trial (see also SI\ref{suppl:when_do_fish_follow}). Consequently, the difference between competent and fixed mode mainly pertained to the number of successful leadership episodes in response to the first few approaches (Figure \ref{fig:sum_follow_durations}). 

\subsection*{Socially competent robots are more efficient}
The number of approaches the robot initiated in a trial was significantly lower for the competent compared to the non-competent agents in experiment 1 (fixed: 26 [4 39], competent: 7 [1 36], U=339.5, P=.0028, CLES=.76) and experiment 2 (random:23 [3 47], competent:14.5 [1 38], U=270, P=.052, CLES=.67).

\begin{figure}[htbp]
    \includegraphics[width=\textwidth]{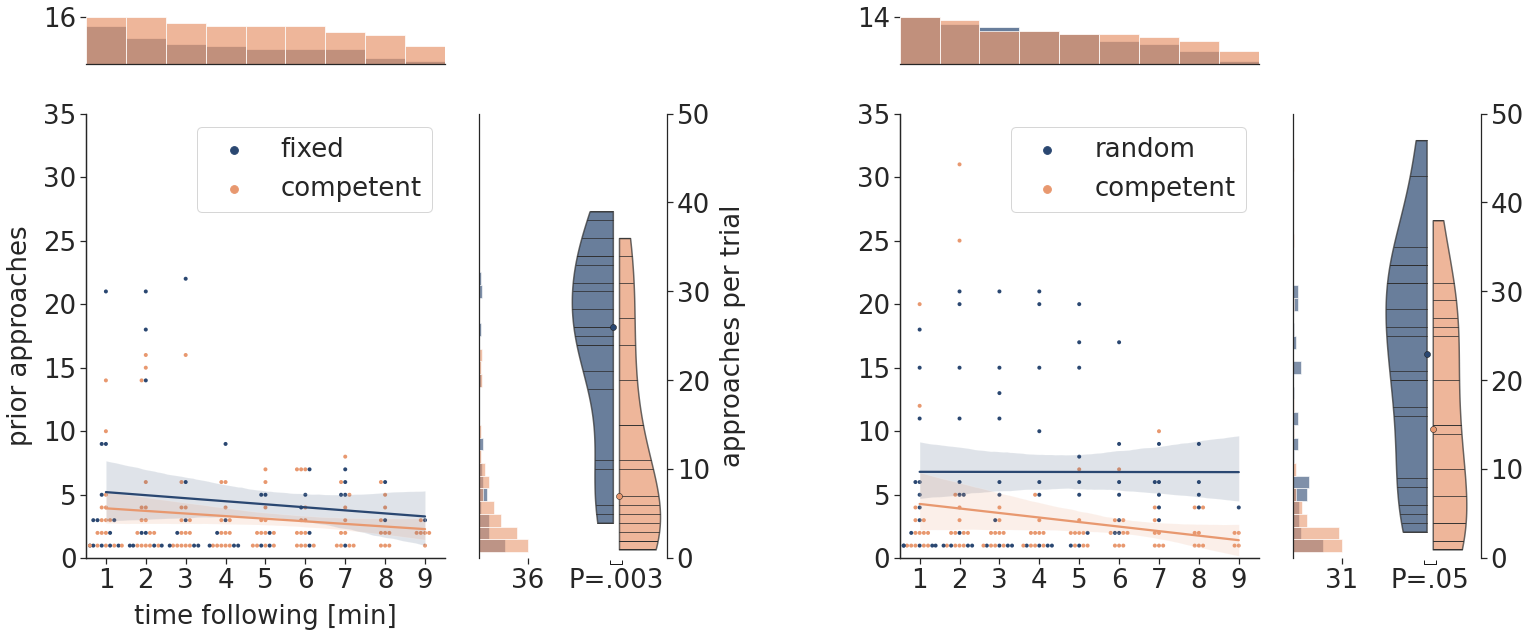}
    \caption{Comparison of approach efficiency. We compared the number of approaches a robot required to elicit a following response for both experiments (left panel: E1, right panel: E2). We determined for each trial how many approaches the robot performed prior to eliciting a following response that in total lasted at least as long as the duration given on the x-axis. Top distributions show the number of animals that were registered for each bracket of following duration. Right marginals show the distributions of approach counts. The the right of each main panel we depict the distribution of the total number of approaches per trial and below the P-value of a Mann-Whitney U-test statistic. In general and for any given duration of a follow episode, socially competent robots require fewer approaches than non-competent robots. }
    \label{fig:num_approaches_follow_durations}
\end{figure}

We then asked how many approaches the robot required for a given total duration of the subsequent follow episodes. In both experiments, the non-competent robot performed more approaches for any given duration of follow episodes (for details see Figure \ref{fig:num_approaches_follow_durations}).

Comparing the two linear regression models in Figure \ref{fig:num_approaches_follow_durations}, the difference between socially competent and random modes appears more pronounced for longer follow episodes. Fewer approaches could indicate longer approach durations; however, we found that our data does not support that view for both, experiment 1 (fixed: 9,5 s [3.1 s 27.1 s], competent: 7.1 s [3.4 s 34.5 s], U=256, P=.38, CLES=.58) and experiment 2 (random: 9.3 s [6.4 s 30.6 s], competent: 7.8 s [3.6 s 26.1 s], U=244, P=.22, CLES=.62).

Naturally, short follow episodes were frequent in both experiments and both respective treatments. Trials with long total follow episode durations (> 6 min) consistently appear more frequently in the socially competent mode.


\section*{Discussion}
We implemented a socially responsive robot which in interactions, with live guppies was found to be more effective and efficient in a leadership task than non-competent robots. 

We tested against two non-competent controls, one that always used the same carefulness (fixed mode) and one that samples its carefulness value from a given reference distribution (random mode). 

We found that the socially competent mode performs better than the fixed mode in all metrics. It produces less avoidance behaviors, on average longer follow episodes and it requires fewer approaches to elicit following behavior. The lower avoidance levels, however, could have been caused by lower motion speeds. The fixed mode was designed to reproduce the mean carefulness of the socially competent mode as measured in pre-trails and it succeeded in doing so. The velocities of the socially competent mode depend, however, on the avoidance behavior of the fish, and the way we selected experimental fish from our holding tank may have introduced a size bias (smaller and therefore younger fish in later trials) which may explain higher carefulness and lower speeds in the socially competent mode. 

In contrast to this finding, the motion speeds of both robot and fish did not differ between treatments in experiment 2. The carefulness values of the socially competent robot were slightly lower, and the fish's avoidance levels even slightly higher than in the random control. We, hence, could not confirm our initial hypothesis of social competence reducing avoidance reactions.
Still, the socially competent robot elicited longer mean follow episode durations using fewer approaches. The differences we observed are less pronounced compared to experiment 1, due to both the socially competent robot being slightly less effective and the random mode being more effective as the fixed mode. A possible explanation for the latter is that the random mode exhibited higher behavioral variability than the fixed mode which may have had an attractive effect. At high carefulness values the robot barely approached the fish. Subsequent follow episodes, hence, are likely caused by the fish coming sufficiently close to the robot on its own.


We reexamined the data of the random treatment in which the robot may still have accidentally changed its carefulness in coherence with our definition of social competence. We found that fish which show increasing avoidance in a given approach phase predominantly follow in the next lead phase if the robot accidentally increased its carefulness (see SI\ref{suppl:accidental_soc_comp}). Due to the randomness of its carefulness choice, the robot, however, is much less predictable from the fish's perspective. It remains to be studied how attractive both high predictability and behavioral variation are in a similar experimental setup.  

To understand better which carefulness values, and therefore which approach strategies, perform well, we implemented a second adaptive behavior that inversely updated the carefulness value: avoidance reactions, thus, lead to bolder robot behaviors and an interest in the robot lead to more careful subsequent approaches. Generally, the majority of live fish were attracted by RoboFish, indicated by a decrease in avoidance scores, especially during the first few approaches. This initial dynamic caused the inverse mode robot to become more careful, reinforcing the same dynamic. Although mathematically possible, we rarely observed the robot become more aggressive as a result of avoidance motions by the live fish, because the defensive behavior of RoboFish pushed the system dynamics into a response region that it could not escape from. At this setting, the robot was barely approaching the fish and did so only very slowly. Hence, the inverse mode effectively simulated a shy and intimidated fish in approach mode, while in lead mode it appeared relatively bold. As we expected, we observed a significant difference in avoidance scores between socially competent and inverse mode. Intriguingly, the inverse strategy resulted in similar leadership performance (see SI).

While the inverse mode appeared to present an unnatural combination of behaviors, and was excluded because we could not control for factors such as motion speed, this result hints at possible future use cases for robots to study social responses to rare or unexpected behaviors.

In summary, we observe long follow episode with both very careful and very bold approaches. It remains unclear why both strategies worked to a similar extent. Live fish may accept leaders with either strategy similarly, or each leader’s strategy could be effective with only a certain subset of the tested population. Sticklebacks prefer to follow individuals whose personality matches their own (Nakayama et al. 2016 Biology Letters) and our previous research found guppies to differ consistently in their following tendencies towards both a robotic leader and another live fish (Bierbach et al. 2018). Thus possible future research might repeatedly test the same individuals for their responses towards different adaptive robotic behaviors.

Live fish across experiments and treatments showed low avoidance reactions towards RoboFish and every cohort included fish that followed the robot closely for several minutes even in the non-competent settings. Biomimetic robots have been increasingly used to study social behavior in species of small freshwater fish \cite{romano_review_2019} and our current results support the feasibility of this approach. 

In an earlier work we proposed that the social acceptance of biomimetic robots might be achieved not only through a realistic reproduction of static and dynamic cues (e.g. visual appearance and motion patterns), but also through implementing probable social conventions, e.g. by matching the robot's response to behaviors that may be expected by interaction partners \cite{landgraf_robofish_2016}. Although the exact mechanism remains unknown, we provide evidence that adaptive, short-term responses may play a crucial role in the ability of interactive biomimetic robots to lead live fish. 

Here, we used avoidance motions as behavioral feedback. Much more complex adaptive rules are conceivable that may use avoidance or other behavioral metrics. Most biomimetic robots, however, have been used in open loop, executing behaviors without feedback from the environment. Incorporating the animals in the control loop of interactive robots allows more complex investigations of the social group dynamics. Almost all interactive robots for the study of animal behavior still use a fixed behavioral policy, i.e. a behavior that always performs the same action when given the same input. Here, we propose the first example of adaptive interactive robots that may be used in studies specifically investigating social responsiveness (see \cite{landgraf_animal---loop_2021} for definitions). 

The ubiquitous presence of fission-fusion societies \cite{aureli_fissionfusion_2008,couzin_behavioral_2006,krause_living_2002} in the animal kingdom highlights that subjects are often approached by familiar or unfamiliar conspecifics. Our behavioral model represents a first example of how observations of the social environment can inform behavioral changes of an adaptive robotic agent. The short-term memory variable used to control the socially-competent agent was designed to mimic the response of a live leader. Our results help demonstrate the importance of social competence and responding to an interaction partner's behavior appropriately to enhance social interactions \cite{taborsky_social_2012,wolf_adaptive_2013}. This work furthermore provides evidence for the feasibility of more complex interaction models of biomimetic robots which have matured into powerful tools for the study of social interactions in animal groups.

\section*{Methods}\label{sec:methods}
\subsection*{RoboFish Setup}\label{sec:robofish_design}
The RoboFish system consists of a glass tank ($120$ cm $\times 120$ cm) that is filled with 7 cm of aged tap water. Four plastic walls separate an experimental area of $100$ cm $\times 100$ cm in the center of the tank. The tank sits on an aluminum rack 1.40 m off the ground. Below the floor of the tank, we operate a two-wheeled differential drive robot on a transparent plastic pane (Figure~\ref{fig:robofish_design}). This robot carries a neodymium magnet directed upwards toward the bottom side of the water tank. A 3D-printed fish replica (Figure~\ref{fig:robofish_design}) is attached to a magnetic base inside the fish tank. This magnet aligns with the robot’s coordinate system. Hence, the replica can be controlled directly by moving the robot. Three red-light LEDs are integrated in the bottom side of the robot. A camera (Basler acA1300-200um, 1280 px $\times$ 1024 px) on the floor faces upwards to localize and track the robot. A second camera (Basler acA2040-90uc, 2040 px $\times$ 2040 px) is fixed 1.5 m above the tank to track both, live fish and replica. The entire system is enclosed in an opaque canvas to minimize exposure to external disturbances. The tank is illuminated from above with artificial LED lights reproducing the daylight spectrum. One personal computer (i7-6800K, 64GB RAM, GTX1060) is used for system operation. A custom robot controller software is used to track the robot in the bottom camera’s feed and control the robot via a Wifi connection. A second program, BioTracker \cite{monck_biotracker_2018}, records the video feed from the top camera, detects and tracks all agents in the tank and sends positional data to the robot control software. For each time step (@25 Hz), the robot control software updates positions and orientations of fish and robot in an internal data structure. Behavior modules can access this object and calculate target positions for the robot as a function of the state currently (or previously) observed. After receiving a new target position from the active behavior, the robot drives towards that target by first rotating and then moving forward with a maximum speed of 30 $cm^{-s}$. All behaviors implemented for this study rely on positional feedback to recruit the fish. Following behavior rarely happens over large distances, hence we implemented variants of a two-staged behavior: the robot first approaches the fish, and then leads it to a target location. For more detailed information on RoboFish operation modes and construction, see \cite{landgraf_robofish_2016}. 
A 3-D printed triangular retainer (“start box”, 19 cm side length) was used to house the fish before the start of the experiment (Figure~\ref{fig:robofish_design}). The retainer contained a cylindrical region with a diameter of 10 cm from which the fish could enter the experimental area through a 3 cm $\times$ 2.5 cm door. Besides the retainer, the environment was otherwise symmetric and monotone. A triangular plastic pane, not shown in Figure\ref{fig:robofish_design}, covered the start box.

\subsection*{Experimental Procedures}
We used identical protocols except for varying the approach behavior as described for each of our three experiments. For each trial, we randomly caught a female guppy from its holding tank and carefully introduced her into the startbox (Figure~\ref{fig:robofish_design}). After one minute of acclimatization, the front door of the startbox was opened. Until the fish left the refuge, RoboFish was set to execute a circular milling movement in front of the refuge's entrance with a diameter of 20 cm and a speed of 8 cm/s. This milling behavior was performed in all experiments to initially attract the live fish as it could see RoboFish from inside the box. Experiments with different behaviors were started as soon as the fish left the startbox (full body length out of shelter). If the test fish did not leave after three minutes, we removed the lid covering the startbox and, after another three minutes the start box was removed entirely. In all experiments, we alternated between modes (socially competent vs fixed/random/inverse) and each live fish was tested only once. Body sizes were measured at the end of a trial to the nearest millimeter and test fish were put back into a holding tank. 

\begin{figure}[tb]
    \begin{floatrow}
        \ffigbox[\FBwidth]{\caption{Illustration of the computation of the next motion target for RoboFish in the approach state. The robot is depicted in blue, the fish in orange.}\label{fig:approach_motion_target}}
        {\centering\includegraphics[width=.55\textwidth]{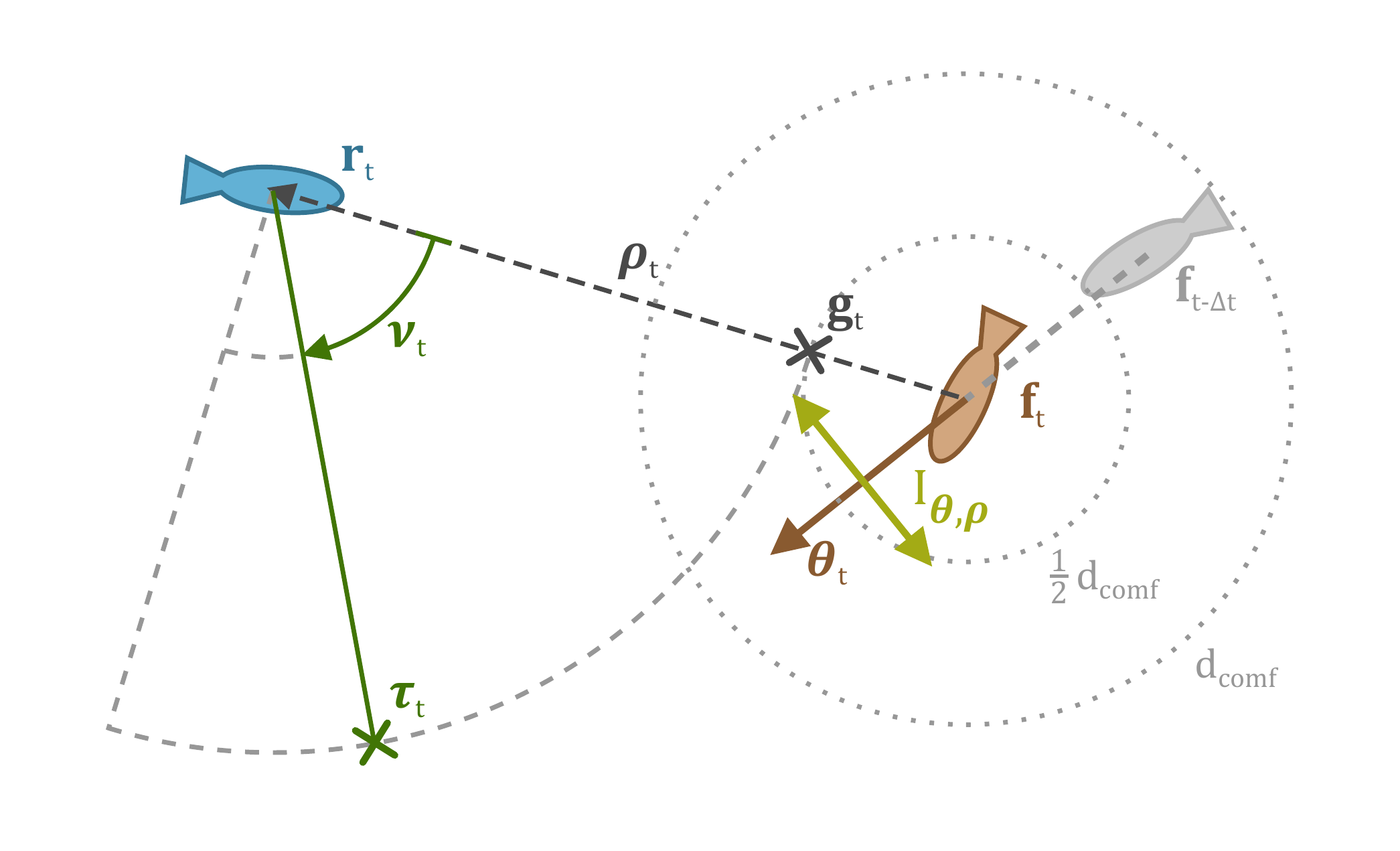}}
        \ffigbox[\Xhsize]{\caption{Illustration of an avoidance event with a negative approach distance. The robot position is denoted in blue, the fish position in orange.}\label{fig:approach_distance}}
        {\centering\includegraphics[width=.4\textwidth]{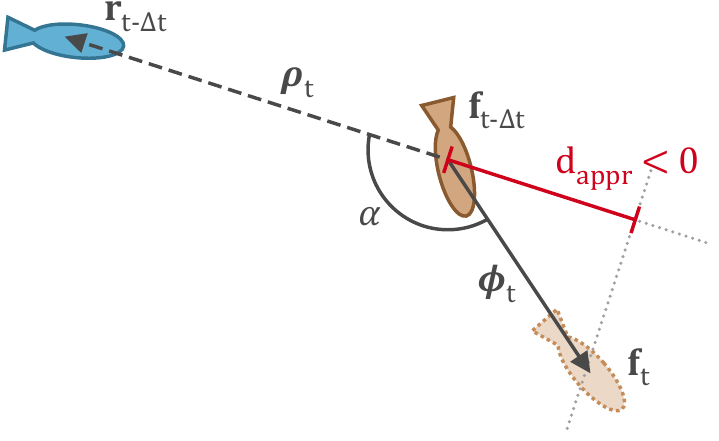}}
    \end{floatrow}
\end{figure}

\subsection*{Quantifying Avoidance and Determining Robotic Carefulness}
In social competence mode, the directness and the speed of the approach are controlled with the carefulness variable $a_{t}$ which represents the robot's memory of the fish's past avoidance responses.
We quantify the avoidance response of the live fish by projecting the motion vector of the live fish during the last time step onto the unit vector between fish and robot. We call this quantity \emph{approach distance}. Given the previous position of the robot \(\vec r_{t-\Delta t}\) and the previous and current position of the fish \(\vec f_{t-\Delta t}\) and \(\vec f_t\), the approach distance can be computed as the inner product of the fish-movement vector $\vec \phi_t = \vec f_t - \vec f_{t-\Delta t}$ with the normalized fish-robot vector $\vec \rho_t = \vec r_{t-\Delta t} - \vec f_{t-\Delta t}$ as
\begin{align}
    d_{\textrm{t}} = \frac{\vec\phi_t^\textrm T \vec\rho_t}{|\vec \rho_t|}.
\end{align}
An illustration of the computation of the approach distance is given in Figure~\ref{fig:approach_distance}. If the approach distance is negative, we consider the live fish to avoid RoboFish and integrate this value into the carefulness variable. The procedure is outlined in the following three steps. 

\begin{enumerate}
    \item \textbf{Clip and normalize negative approach distances}
    

\begin{align}
    e_t =
    \begin{cases}
       \left|-d_t\right|_{v_s}^{v_p} & \text{if $d_t < 0$} \\
        0 & \text{otherwise}
    \end{cases}.
\end{align}

The notation \(|\cdot|_a^b\) refers to clipping the value to the range \([a,b]\) and then normalizing to \([0, 1]\). For our experiments, the bounds were empirically determined:  $v_s = 2.5$ and $v_p = 10$. Hence, slow or tangential movements are mapped to $0$, fast movements away from the robot are mapped to $1$. 

\item \textbf{Disregard when far away and exponential smoothing}\\
Avoidance movements at the other end of the tank may not relate to the robot's behavior. We hence disregard fish motions outside an assumed interaction zone $d_\textrm{I} = 56$ cm.  
We compute the \textit{avoidance score}  $\bar e_t$ as an exponential average of the normalized negative approach distances. We initialize $\bar e_0 = 0.5$ at the beginning of each trial and update as
\begin{align}\label{eq:avoidance_score}
    \bar e_t = \left|\big(\beta I_s s_e e_t + (1 - \beta) \bar e_{t - 1}\big)\right|_{0.0}^{1.0}.
\end{align}
Here, $\beta = 0.0025$ is the smoothing factor, \(I_s\) is an indicator that is set to 1 if the fish is within the interaction zone and 0 otherwise, and \(s_e = 8.0\) scales the incoming avoidance responses.
    
\item \textbf{Calculation of carefulness}\\
We incorporate the \textit{avoidance score} relative to the baseline $b_e = 0.5$ into the \textit{carefulness variable} again as an exponential average: 
\begin{align}
    a_t = \big|(1-\eta)a_{t-1} + \eta(\bar e_t - b_e) \Delta t \big|_{0.0}^{1.0},
    \label{eq:carefulness}
\end{align}
where \(\Delta t\) is the time step and \(\eta = 0.075\) is another smoothing factor. The carefulness variable, hence, is increased if the avoidance score is above the baseline, and decreased otherwise.
\end{enumerate}

The robot's next target location is a function of the carefulness variable. We first calculate the default target $\vec g_t$, $6$ cm away from the fish, on the line connecting robot and fish.
The robot then rotates $g$ around its position proportional to its current carefulness $a_t$:
\begin{align}
    \vec \tau_t = \vec R(\nu_t) (\vec g_t - \vec r_t) + \vec r_t,
\end{align}
with rotation matrix \(\vec R(\nu_t)\) and the rotation angle \(\nu_t\) as a function of the approach parameter \(a_t\) as
\begin{align}
    \nu_t = a_t \frac{1}{2} \pi I_{\theta,\rho}.
\end{align}
Here, \(I_{\theta,\rho}\) is an indicator which is positive if the robot is left of the fish (w.r.t. its movement direction \(\vec \theta_t\)), and negative otherwise. This makes careful approaches turn into the movement direction of the fish. 
The carefulness variable scales the approach angle up to 90\textdegree{} such that maximally careful robots move perpendicular to $\vec \rho_t$ at $a_t = 1$, circling around the fish (see Figure~\ref{fig:approach_motion_target})

The carefulness variable also affects the robot's movement speed through a scaling factor \(s_t = 1 - a_t + s_c\) where \(s_c = 0.2\) is its base speed. Hence, the maximum forward speeds are reached with $a_t = 0$ and $s_t = 1.2$. A set of low-level PID controllers is used to to calculate the motor speeds for turning towards and approaching the target. Linear ramps are used for smooth acceleration and stopping at target arrival. Note that due to the motion of the live fish and the high update rate, a new target point is computed before the robot reaches the previous one in virtually all cases. 

Once the fish has been approached and it stays within a \(d_\textrm{comf} = 12\) cm distance for more than $2$ s, the behavior switches to \emph{lead phase} unless the fish is too close \(d < 6\) cm, in which case the behavior remains in the approach mode until the fish is back within the robot's comfort zone.

In \emph{lead phase}, the robot tries to lead the fish along the walls of the tank. We define points close to the corners of the tank with a distance of 10 cm to the two adjacent walls as target points and cycle through these points clock-wise to select the next target.
The robot does not drive to each target in one continuous pass but rather in short motion bursts using a sequence of target points. Each subsequent target location is calculated as a point 15 cm away on the line between robot and corner or the corner itself, if the robot is sufficiently close. Before the robot continues to its next target, it waits until the fish is within a distance of 28 cm. If fish and robot are farther apart for more than one second, the robot switches back to approach phase.
During lead phase, the robot moves with a speed factor $s_t = 0.8717$.


\subsection*{Pretrials}\label{sec:suppl_pretrials}
Prior to the main experiments, pretrials were conducted in competent mode with $N=20$ single fish. The carefulness variable was initialized to $a_t=0.5$. All carefulness values were collected throughout all pretrials resulting in the reference distribution (Table \ref{tab:suppl_ref_distribution}) used in the implementation of the random control (experiment 2) and the fixed mode (experiment 1). 

\begin{table}[h!]
    \centering
    \begin{tabular}{cc}
    \textbf{interval} & \textbf{norm. frequency} \\
$[0.0, 0.1]$ & $0.112739$ \\
$(0.1, 0.2]$ & $0.031610$ \\
$(0.2, 0.3]$ & $0.034126$ \\
$(0.3, 0.4]$ & $0.042342$ \\
$(0.4, 0.5]$ & $0.069718$ \\
$(0.5, 0.6]$ & $0.080316$ \\
$(0.6, 0.7]$ & $0.065151$ \\
$(0.7, 0.8]$ & $0.108997$ \\
$(0.8, 0.9]$ & $0.126346$ \\
$(0.9, 1.0]$ & $0.328655$ 
    \end{tabular}
    \caption{Reference distribution for carefulness values. In experiment 2 each approach phase is performed with a value drawn from this reference distribution. Each sample corresponds to the respective bin center.}
    \label{tab:suppl_ref_distribution}
\end{table}

\subsection*{Non-competent and Inverse-Competent Behaviors}
We compare the socially competent mode against three controls: two non-competent and one inverse-competent behavior.

In \emph{fixed mode}, we used the mean carefulness ($\bar{a}=0.528$) in every approach phase. This resulted in a constant approach angle of $\approx47$\textdegree and constant approach speed of $19 cm^{-s}$.

In \emph{random mode}, the robot randomly samples a carefulness value $a_t$ at the start of each approach phase from a target distribution $\chi_{a}$ (initially set to the reference distribution) and uses this value throughout the approach phase. Before we sample a new $a_t$ in the subsequent approach phase, we correct $\chi_{a}$ by subtracting the respective proportion of time the robot was in the last approach mode from the respective bin. This allows matching the reference distribution approximately.

In \emph{inverse mode}, we flipped a sign in the calculation of the carefulness variable (Equation \ref{eq:carefulness}) such that above-threshold avoidance scores lead to a reduction, and below-threshold avoidance scores lead to an increase of the carefulness variable.



\subsection*{Quantifying Leadership Performance: the "Follow" Metric}\label{sec:methods_follow}
Similar to the avoidance response, we measure following behavior by projecting the motion vector of the live fish during the last time step onto the unit vector between fish and RoboFish. Given the position of RoboFish at the previous time step \(\vec r_{t-\Delta t}\) and the fish's position at the previous time step \(\vec f_{t-\Delta t}\) and at the current time step \(\vec f_t\), this value can be computed by taking the inner product of the fish-movement vector $\phi_t = \vec f_t - \vec f_{t-\Delta t}$ with the normalized fish-robot vector $\rho_t = \vec r_{t-\Delta t} - \vec f_{t-\Delta t}$ as
\begin{align}
    d_{\textrm{t}} = \frac{\vec\phi_t^\textrm T \vec\rho_t}{|\vec \rho_t|}.
\end{align}
If this value is positive, i.e., if the projected movement vector points towards the robot, we consider it as evidence for attraction which we compute as 
\begin{align}
    o_t =
    \begin{cases}
       \left|d_t\right|_{v_s}^{v_p} & \text{if $d_t > 0$} \\
        0 & \text{otherwise}
    \end{cases},
\end{align}
where we use the notation \(|\cdot|_a^b\) to denote that the value is clipped to the range \([a,b]\) and then normalized to \([0, 1]\).
Here, the lower bound $v_s$ and the upper bound $v_p$ are hyper-parameters of the algorithm and were empirically determined, analogously to the computation of the avoidance score: $v_s = 2.5$ and $v_p = 10$.
If the projected movement is negative, we consider it an avoidance (see main text). The follow score is computed similarly to the avoidance score (see main text). It is initialized to 0.5 at the beginning of each trial and then updated at each time step with the follow events $o_t$ as
\begin{align}
    \bar o_t = \left|\big(\beta_o I_s s_o c_o o_t + (1 - \beta_o) \bar o_{t - 1}\big)\right|_{0.0}^{1.0},
\end{align}
where \(c_o\) is a correction term defined as
\(c_o = 1 + \exp\left(-\frac{1}{3}o_t\right)\), $\beta_o = 0.005$ is the learning rate, and $s_o = 2.0$ scales the follow event.

\begin{figure}[htbp]
    \centering
    \label{fig:true_lead}
    \includegraphics[width=1\textwidth]{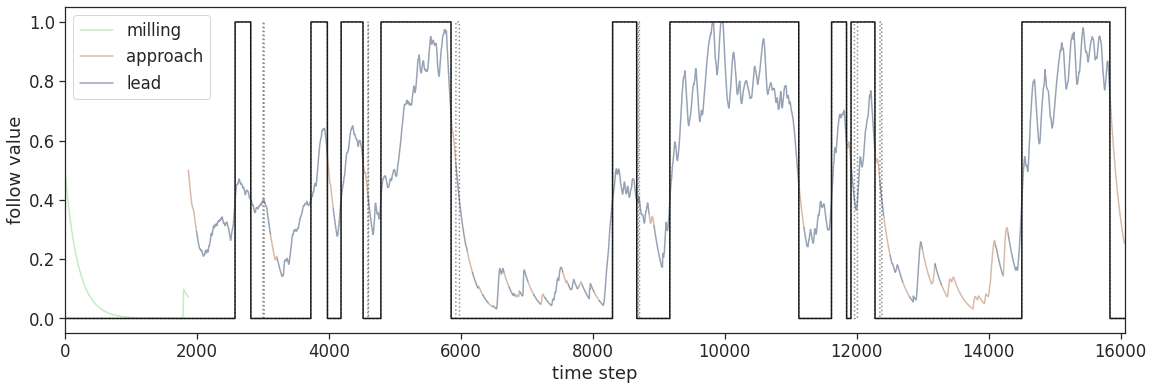}
    \caption{Binarization of follow value into episodes of following behavior. The follow metric over time is depicted here for the different behavioral phases of the robot, i.e. the milling phase (green), approach phases (red) and lead phases (blue). The dashed black line depicts instances with above-threshold follow values that were removed. The result is depicted with a solid black line, the following duration hence is the width of each of these blocks.}
    \label{fig:follow_calculation}
\end{figure}

In contrast to the duration of the robot's lead phase, the follow metric more accurately reflects whether the fish was actually following. In fact, the robot much more often switches engages in a (short) lead phase than fish actually show noticeable follow episodes. 
We define that duration as an episode in which RoboFish is in its lead phase and the follow value $\bar o_t$ is above a threshold of 0.4.
We bridge small (less than 200 time steps wide) gaps between follow episodes and remove remaining short episodes of following behavior (less than 200 time steps) by applying erosion and dilation operations. An example visualizing this process is shown in Figure \ref{fig:follow_calculation}

\subsection*{Test Fish and their Maintenance}
\label{sec:behavioral_experimentation}
We used Trinidadian guppies (Poecilia reticulata) that are descendants of wild-caught fish from the Arima-River system in Northern Trinidad. Test fish came from large, randomly outbred single-species stocks maintained at the animal care facilities at the Department of Life Sciences, Humboldt University of Berlin. To avoid inbreeding, stocks are regularly supplemented with wild-caught animals brought back from fieldwork in Trinidad and Tobago. We provided a natural 12:12h light:dark regime and maintained water temperature at 25\textdegree C. Fish were fed twice daily ad libitum with commercially available flake food (TetraMin™). For the experiments, only female guppies were used to avoid effects of sex-specific differences in responsiveness.

\subsection*{Statistical analysis}
We used the conservative Mann-Whitney U tests to compare average behavioral measures between treatments and Student-t tests to compare variables at different approaches. All analyses were performed using Python.

\subsection*{Ethics note}
Experiments reported in this study were carried out in accordance with the recommendations of “Guidelines for the treatment of animals in behavioural research and teaching” (published in Animal Behavior 1997) and comply with current German law approved by LaGeSo Berlin (G0117/16 to D.B.).

\bibliographystyle{plain}
\bibliography{references}

\appendix
\section*{Supplementary Information}



\subsection{Body size of test fish}\label{sec:suppl_body_size}
Body size of all test fish was measured as Standard Length (from tip of snout to end of caudal peduncle) by transferring the fish into a Petri dish filled with water and placed above millimeter paper.  Digital photographs of the individual fish were taken and body size measured using ImageJ \cite{rueden2017imagej2}.

Experiment 1: The body size of fish tested with socially competent RoboFish was SL 28.4 mm ($\pm$4.1 SD) and for those tested with non-competent (fixed carefulness) RoboFish SL 29.5 mm ($\pm$ 5.4 SD). There was no significant difference in body size between both tested cohorts (unpaired \textit{t}-test: $t_{44}=0.83$; $P=0.41$). 
Experiment 2: The body size of fish tested with socially competent RoboFish was SL 31.2 mm ($\pm$5.5 SD) and for those tested with non-competent (random carefulness) RoboFish SL 30.1 mm ($\pm$ 4.2 SD). There was no significant difference in body size between both tested cohorts (unpaired \textit{t}-test: $t_{40}=0.74$; $P=0.47$). 
Experiment 3: The body size of fish tested with socially competent RoboFish was SL 28.11 mm ($\pm$3.9 SD) and for those tested with inverse-competent RoboFish SL 28.76 mm ($\pm$ 3.6 SD). There was no significant difference in body size between both tested cohorts (unpaired \textit{t}-test: $t_{34}=0.49$; $P=0.62$). 

\subsection{Interaction strength over inter-individual distance}
\label{sec:follow_iid}
The robotic interaction behavior was modeled with two distinct phases, "approach" and "lead". This design decision was motivated by a preceding analysis of trajectory data of two live Guppys \cite{maxeiner_imitation_2019}. To assess at which inter-individual distances live fish show following behavior we calculated the projection of the focal agent's motion vector onto the unit vector pointing to its interaction partner (the "follow" metric, see Methods). The dataset was recorded previously for another experiment. Two fish were randomly selected from a bigger tank, moved to the RoboFish tank and recorded for 10 minutes without disturbance. After the observation period they were put into a third tank to prevent observing the same animals twice. 

\begin{figure}[htbp]
    \centering
    \includegraphics[width=1\textwidth]{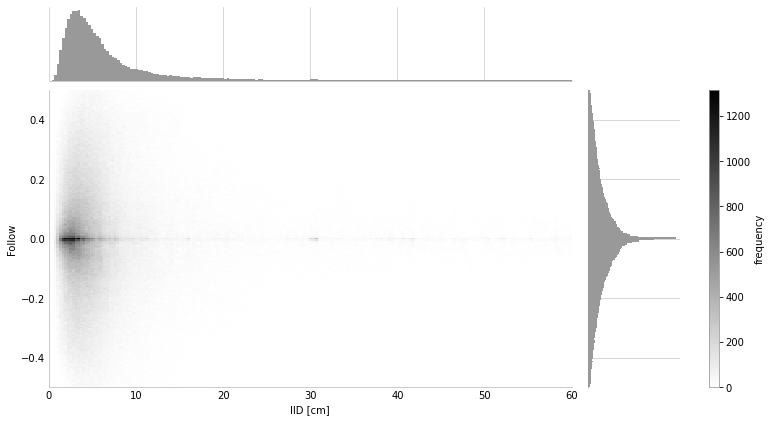}
    \caption{Distribution of follow values over inter-individual distances.}
    \label{fig:distribution_follow_live}
\end{figure}

Most follow values fall into a close interaction range (see pronounced peak at $\approx 3$ cm in Figure \ref{fig:distribution_follow_live}) and fall off in both frequency and magnitude with animals further apart than 9 cm ($\approx3$ body lenghts).





\subsection{When and how frequently do fish follow?}
\label{suppl:when_do_fish_follow}
\begin{figure}[tb]
    \begin{floatrow}
        \ffigbox[\FBwidth]{\caption{Distribution of follow episodes over start time.}\label{fig:following_events_over_starttime}}
        {\centering\includegraphics[width=0.6\textwidth]{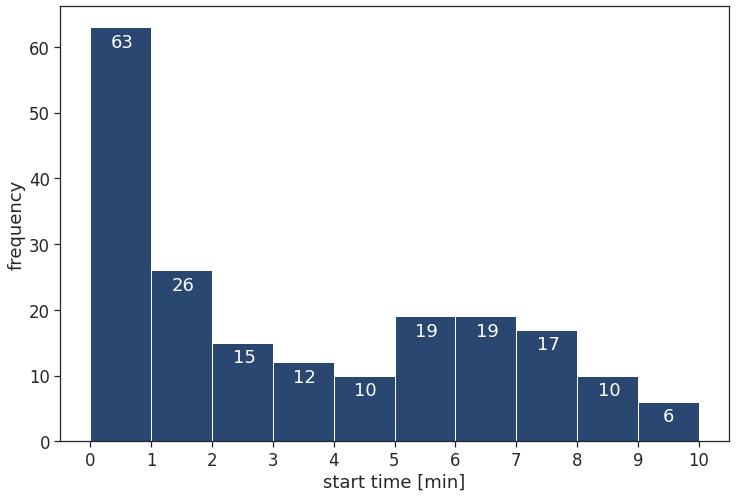}}
        \ffigbox[\Xhsize]{\caption{Duration of follow episodes over the respective start time within the trial.  }\label{fig:following_duration_over_starttime}}
        {\centering\includegraphics[width=0.4\textwidth]{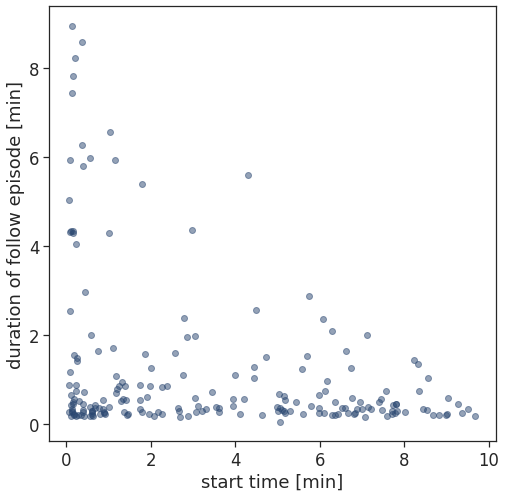}}
    \end{floatrow}
\end{figure}

Episodes of following behavior were determined as described in the methods section and the start times were extracted for all (N = 197) following instances from both experiments. Figure \ref{fig:following_events_over_starttime} shows the respective distribution of follow episodes over their start times. Half of all follow episodes occur within the first three minutes (104 / 197), accounting for 74 \% of the combined following durations (173 min / 235 min). Over all experiments we recorded a total of 3.9 hours of following behavior which corresponds to 27 \% of the total duration of all 86 trials (see Figure \ref{fig:following_duration_over_starttime} for duration of follow episodes over start time).

\subsection{Relation of carefulness and motion speed}
\label{suppl:speed_vs_carefulness}
\begin{figure}[htbp]
    \centering
    \includegraphics[width=0.75\textwidth]{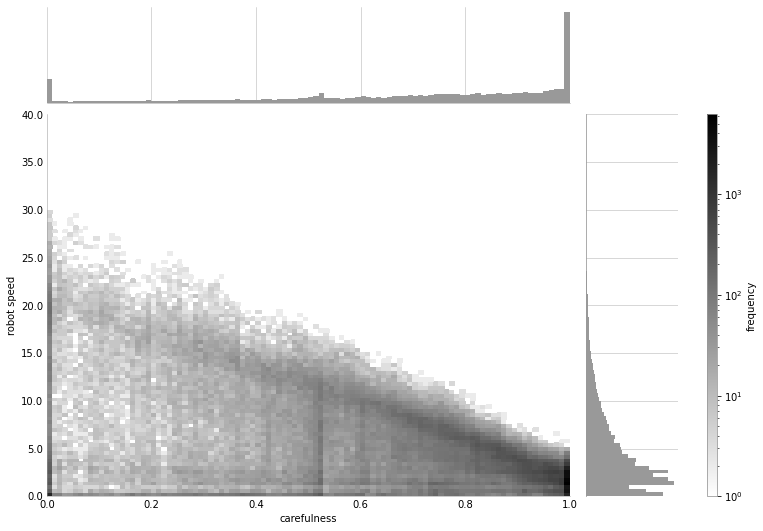}
    \caption{Motion speed of the robot over the carefulness variable. Since the robot moves in bursts, accelerating from and decelerating before arriving at target locations, all motion speeds up to a maximum are possible. As described in the Methods section, this maximum is a linear function of the carefulness variable with negative slope.}
    \label{fig:speed_vs_carefulness}
\end{figure}

There is a direct relationship of approach motion speed and the carefulness variable. We have extracted the motion speeds for all trials in social competent mode (from experiments 1-3) and visualized the dependency in Figure \ref{fig:speed_vs_carefulness}. 

\subsection{Analysis of accidental social competence in random mode}
\label{suppl:accidental_soc_comp}
In random mode (experiment 2), the carefulness value is sampled randomly from a reference distribution. RoboFish may therefore change its carefulness such that it is, by chance, coherent with our definition of social competence. To assess whether it were predominantly those instances responsible for follow episodes, we calculated the average change in the raw avoidance scores (negative approach distances $d_{\textrm{t}}$) throughout an approach phase (termed $\Delta q_{t}$) and relate them to the (random) change of the carefulness variable (termed $\Delta a_t$). 

We recorded following episodes over all values of $\Delta a_t$, i.e. even robots that became bolder ($\Delta a_t < 0$ recruited fish. 

\begin{figure}[htbp]
    \centering
    \includegraphics[width=0.75\textwidth]{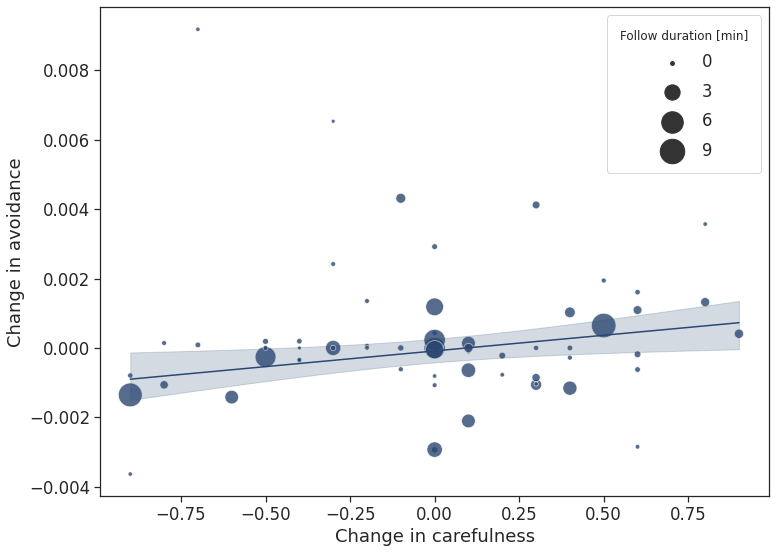}
    \caption{Dependency of follow episodes on the change of the robots carefulness and the change of the animal's mean avoidance.}
    \label{fig:delta_avoid_vs_delta_carefulness}
\end{figure}

We find that lead phases ensuing approach phases with increasing avoidance scores were more successful, i.e. resulted in following behavior, when the robot increased its carefulness, as shown in Figure \ref{fig:delta_avoid_vs_delta_carefulness}. The correlation is not very pronounced ($R^2 = 0.084$) but supports our main results, i.e. that socially competent leaders perform better than non-competent onces. 

\subsection{Comparison of motion speeds}
\label{suppl:speed_comparison}
\begin{figure}[htbp]
    \centering
    \includegraphics[width=\textwidth]{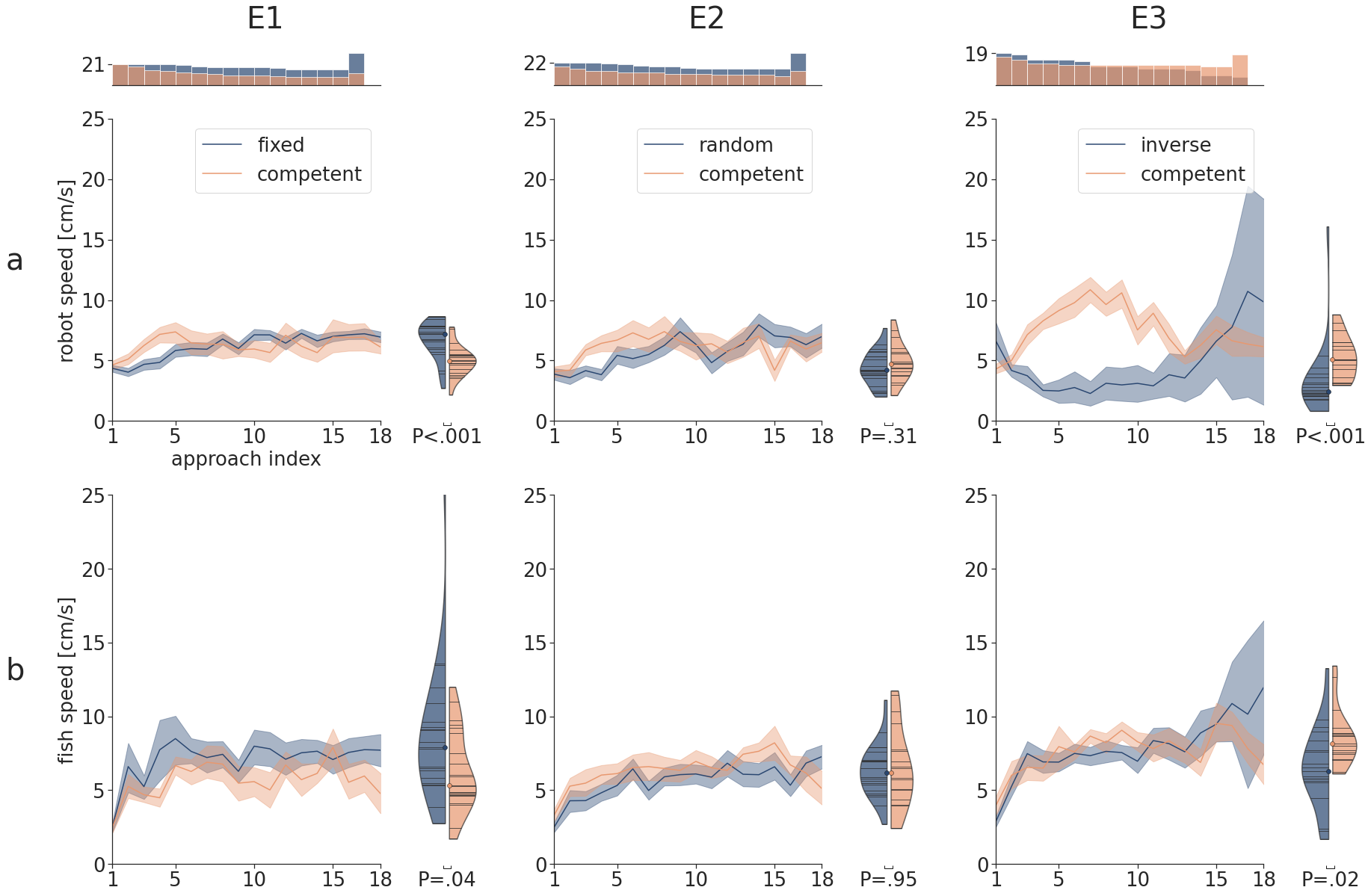}
    \caption{Comparison of motion speeds averaged over all time steps in approach phases for robot (row a) and fish (row b). We compare the socially competent mode (orange) with fixed mode (blue, column E1), random mode (blue, column E2) and inverse mode (blue, column E3). Every panel shows the distribution of motion speeds over the approach index and the per-trial means in a double-violin plot. To detect differences between per-trial speed means, we performed a Mann-Whitney U-test. P-value are given under the violin plots. }
    \label{fig:speeds}
\end{figure}

We asked whether the robot operated at different speeds between treatments which could explain differences in leading performance. We averaged all motion speeds along approach phases and depict the mean speed over all trials and treatment in Figure \ref{fig:speeds}. 
We found that motion speeds of fish in experiment 1 differed significantly between socially competent and fixed mode (U=301, P=.044, CLES=.68) as did motion speeds of the robots (U=361, P<.001, CLES=.82).
In experiment 2, we found that motion speeds of both the fish and the robot did not differ significantly between treatments (fish: U=201, P=.95, CLES=0.51; robot: U=160, P=.31, CLES=.6).
In experiment 3, both speeds differed significantly (robot: U=49,	P<.001, CLES=.85; fish: U=85, P=.016, CLES=.26).

\subsection{Mean follow episode durations}
\label{suppl:mean_follow}
\begin{figure}[htbp]
    \centering
    \includegraphics[width=\textwidth]{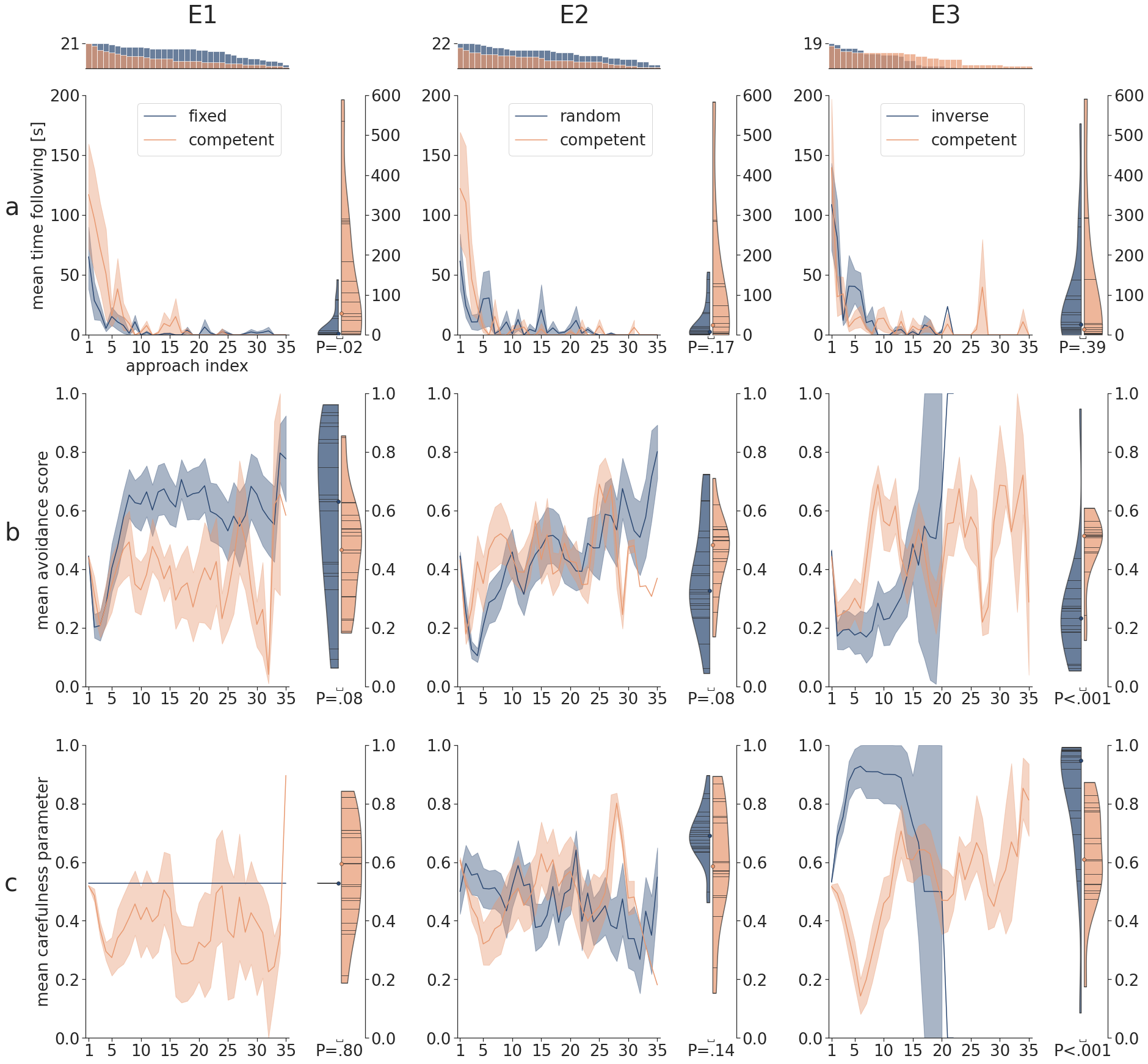}
    \caption{Comparison of follow episode durations. We compare the follow episode durations within a trial for socially competent robots (orange) and non-competent robots (blue) for both main experiments (columns E1 and E2) and an additional experiment in which we tested an inversely competent robot (E3, see Methods). Each panel shows the mean duration of follow episodes over the approach index (i.e. a sequential ID of approach phases, left sub-panel). The approach phase count is shown in the bar plot above the left sub-panels. The distribution of the mean following durations (right sub-panel) is depicted in the half-violin plots. Median values are depicted with an orange and blue circle, respectively. P-values of a Mann-Whitney U-test are given under the violin plot. Long follow episodes are predominantly initiated at the beginning of the trial, after the first few approaches. Differences between treatments pertain to the first 5 approaches in which the socially competent robots perform significantly better.}
    \label{fig:mean_follow_durations}
\end{figure}

We compare the follow episode durations within a trial for socially competent robots and non-competent robots for both main experiments (experiment 1 and experiment 2) and an additional experiment in which we tested an inversely competent robot (experiment 3, see Methods). Figure \ref{fig:mean_follow_durations} shows the mean duration of following episodes over approach indices and per-trial means, and analogously, the avoidance and carefulness distributions in all experiments.   

\subsection{How do fish move relative to the robot?}
\label{suppl:cumsum_projections}
\begin{figure}[htbp]
    \centering
    \includegraphics[width=\textwidth]{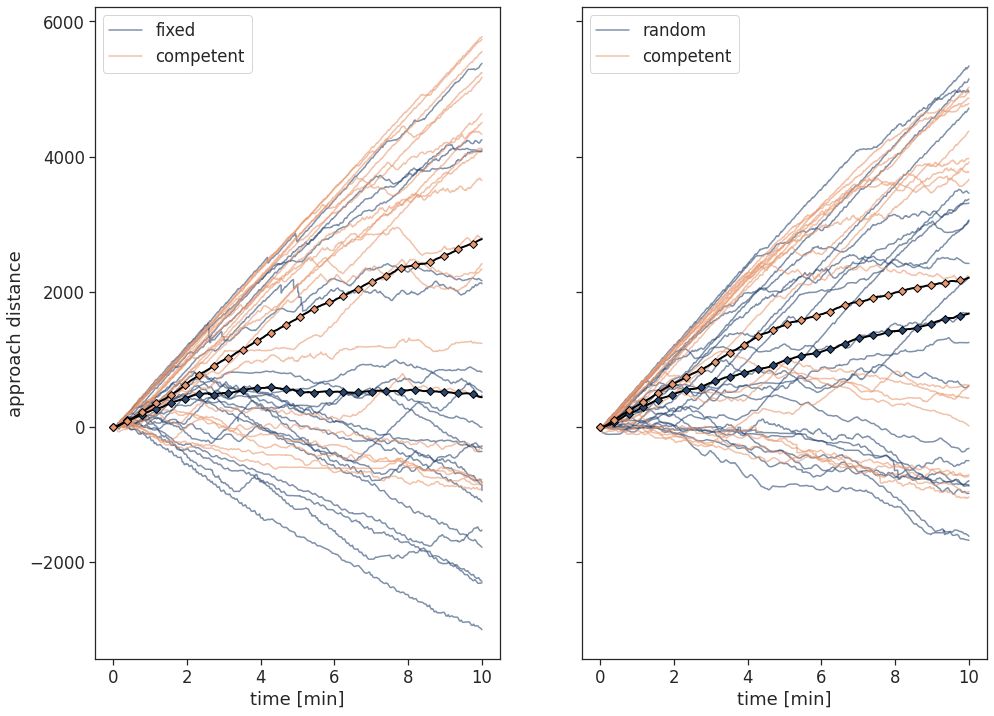}
    \caption{Raw motion vectors projected onto the direction of the robot from the perspective of the fish. Calculating the cumulative sum expresses how much the fish was avoiding or approaching the robot. The two panels show individual trials for the socially competent mode (orange lines) and the non-competent controls (blue lines. In experiment 1 (left panel) the average motion (bold lines) is positive (towards the robot) for both treatments only in the first few minutes of the trial. After that, fish that interacted with the fixed mode moved on average such that avoidance and attraction balanced each other, in contrast to the socially competent mode in which attraction dominated until the end of the trial. In experiment 2, both averages are closer together, with a higher slope for the socially competent mode for approximately the first third of the trial. The random mode is more attractive than the fixed moded as can be seen by an almost constant positive slope.}
    \label{fig:cumsum_raw_projections}
\end{figure}

Projecting the motion vector of the fish onto the unit vector that points to the robot's location (a quantity we termed \textit{approach distance}, see Figure \ref{fig:approach_distance}), we can quantify how strongly the fish is attracted (positive values) and repelled by the robot (negative values). 
We visualized these motion projections as cumulative sum over time for each tested animal in figure \ref{fig:cumsum_raw_projections}. Socially competent robots differ from fixed and random controls in that they on average show more positive values throughout the entire trial. 

\end{document}